\UseRawInputEncoding
\documentclass{amsart}

\usepackage{enumerate}
\usepackage{amsmath}
\usepackage{lmodern}
\usepackage{listings}
\usepackage[rmargin=4cm,lmargin=4cm]{geometry}
\usepackage[dvipsnames]{xcolor}
\usepackage{alltt}
\usepackage{fancyvrb}
\usepackage[colorinlistoftodos,prependcaption,textsize=tiny]{todonotes}
\usepackage[T1]{fontenc}
\usepackage{hyperref}
\usepackage{tikz}

\usepackage{ifthen}
\usepackage{graphicx}
\usepackage{amssymb}
\usepackage{enumerate}
\usepackage{url}
\newtheorem{anyprop}{Anyprop}[section]

\newtheorem{theorem}[anyprop]{Theorem}

\theoremstyle{definition}

\begin{document}
\title[Artificial Conceptual Generation of Topological Groups]
{Artificial Cognitively-inspired Generation  of the Notion of Topological Group in the Context of Artificial Mathematical Intelligence}


\author[Danny A. J. G\'omez-Ram\'irez]{Danny A. J. G\'omez-Ram\'irez}
\author[Yoe A. Herrera]{Yoe A. Herrera-Jaramillo}
\author[Florian Geismann]{Florian Geismann}
\address{ Parque Tech at the Instituci\'on Universitaria Pascual Bravo, Medellín, and Visi\'on Real Cognitiva S. A. S. Itagüí, Antioquia, Colombia.}
\address{Instituci\'on Universitaria Tecnol\'oligo de Antioquia, Medell\'in, Colombia.}
\address{Institute of Discrete Mathematics and Geometry, Vienna University of Technology
Vienna, Austria.}
\email{daj.gomezramirez@gmail.com}
\email{yherrera743@unab.edu.co}
\email{fgeismann@live.at}



\begin{abstract}
The new computational paradigm of conceptual computation has been introduced in the research program of Artificial Mathematical Intelligence. We provide the explicit artificial generation (or conceptual computation) for the fundamental mathematical notion of topological groups. Specifically, we start with two basic notions belonging to topology and abstract algebra, and we describe recursively formal specifications in the Common Algebraic Specification Language (CASL). The notion of conceptual blending between such conceptual spaces can be materialized computationally in the Heterogeneous Tool Set (HETS). The fundamental notion of topological groups is explicitly generated through three different artificial specifications based on conceptual blending and conceptual identification, starting with the concepts of continuous functions and mathematical groups (described with minimal set-theoretical conditions). This constitutes in additional heuristic evidence for the third pillar of Artificial Mathematical Intelligence.

\end{abstract}

\maketitle


\smallskip


\section{Introduction}
Latest advances in computational creativity, cognitive and computer science continue enhancing our understanding about the way in which our minds create mathematics at high levels of sophistication \cite{cinventionbook}. In particular, more precise formalization of fundamental cognitive mechanisms for conceptual creation has been developed and tested in several mathematical domains \cite{martinez2016,conceptinvention1, hdtp,gomezramirezetal2018}. 
Among those basic cognitive abilities conceptual blending has shown to be not only one of the most powerful, but also one of the most omnipresent among mathematics \cite{Alexander,fauconnierturner}. For instance, seminal notions of (algebraic) number theory, Fields and Galois theory and commutative algebra have been conceptually meta-generated (with a computational basis) in terms of a categorical formalization of conceptual blending \cite{conceptinvention1, conceptinvention2, gomezramirezetal2018, gomezetal}.

More concretely, the blend of a `V'-shaped diagram between two input mathematical concepts with a generic (base) concept is characterized by means of categorical colimits. Now, such a colimit exists because mathematical concepts are formalized in terms of many-sorted first-order theories with axiom-preserving morphisms \cite[\S 8.2]{conceptinvention1}. 
Another basic metamathematical cognitive mechanism commonly used during mathematical research is \emph{conceptual identification}; i.e., the ability of (cognitively) interpreting two (abstract) concepts as the same, with the purpose of simplifying inferential processes on the mind \cite{holte1986, kurcz1986}. 



Research in this direction is also closely related with the development of new forms of cognitively-inspired artificial intelligence on the domain of abstract mathematical discovery/creation \cite{conceptinvention1}, specifically, within the multidisciplinary research program called Artificial Mathematical Intelligence (AMI) \cite[\S8.7.1]{conceptinvention1,AMI}.\footnote{www.ArtificialMathematicalIntelligence.com}  
	
	This AMI research meta-project is a multidisciplinary program for generating the necessary (theoretical and computational) foundations for the construction of a universal mathematical artificial agent (UMAA) \cite{AMI}.
	More particularly, artificial mathematical intelligence, or cognitive-computational metamathematics, requires the extension of the classic computational paradigm towards an enhanced multidisciplinary framework for developing \emph{conceptual computation}, with all the (meta)mathematical foundational setting involved \cite{AMI-Ch1}, \cite{AMI-Ch2}.
	Furthermore, the first pillar of this new field of research comprehends the \emph{New Cognitive Foundations' Program}, which involves the development of new and refined (meta)mathematical frameworks and structures like Dathematics (a dual version of mathematics based upon proper classes) and the physical numbers (a multidisciplinary refinement of the natural numbers). And in a broader sense, the (cognitive) reality of mathematics and the mathematics of the (cognitive) reality, among others \cite{AMI-Ch3}, \cite{AMI-Ch4}, \cite{AMI-Ch5} and \cite{AMI-Ch6}.

	The second fundamental pillar is the establishment and formalization of a robust \emph{typology of fundamental cognitive (metamathematical) mechanisms} used by the mind during formal creation \cite{AMI-Ch7}, \cite{AMI-Ch8}, \cite{AMI-Ch9} and \cite{AMI-Ch10}. The pursuit for such a global typology is important from a metamathematical point of view, as well as from a mental and biological perspective. In fact, one of the most antique and relevant open questions in science is the general functioning of the mind. Such a inquiry involves the identification of the fundamental mechanisms used by the mind during theoretical and applied reasoning \cite{AMI-Ch10}. 
	
	The last pillar of the AMI meta-project requires the specific generation of general pseudo-(pre)-code of special forms of conceptual computation sound enough for constructing (gradually) better and more holistic versions of UMAA-s in starting mathematical sub-disciplines, and the most reasonable extensions of the AMI program to further related fields like finances, biology and physics \cite{AMI-Ch11}, \cite{AMI-Ch12}.
	
	Artificial mathematical intelligence has shown outstanding applications even beyond the setting of metamathematics and computer science. In fact, the development of a general typology of fundamental cognitive (metamathematical) mechanisms used in formal creation/invention has allow us to describe cognitive and morpho-syntactic prevention guidelines for COVID-19 \cite{gomez2021prevention} and \cite{gomez2021morpho}.


In the modern literature concerning formal (artificial) conceptual generation based on conceptual blending and metaphorical reasoning etc., an special attention has been set to the study of algebraic and arithmetic notions. On the other hand, concepts with a more topological and geometrical nature represents a relatively unexplored field in this regard. So, we aim to start to fill this gap presenting a detailed conceptual generation of the seminal concept of topological group starting with the elementary notions of group, continuous function (between topological spaces) and `perfect square' topological space. We will present the corresponding pseudo-specifications described with the common algebraic specification language (CASL) \cite{CASLuserguide} and implicitly using the formalisms described by the Heterogeneous Tool Set (HETS) \cite{mossakowskihets}. HETS is a suitable software because it provides specific tools for computing formal colimits for the above formalization of concepts described above. 

\section{Methods}

\subsection{Conceptual Preliminaries}
For the sake of completeness in the presentation, we recall the initial notions that we will use as foundational bricks of our conceptual `building'. 

First of all, a group $(G,+,e)$ is simply a set equipped with a binary operation $+$ and an outstanding element $e\in G$, such that $e$ is the neutral element with respect to $+$, $+$ is associative and each element possesses an inverse. 

Second, a topological space $(X,T)$ consists of a set $X$ and a collection $T$ of subsets of $X$ satisfying the following conditions: $\emptyset,X\in T$, $T$ is closed under finite intersections and arbitrary unions. So, a function $f:X\rightarrow Y$ between topological spaces $(X,T_X)$ and $(Y,T_Y)$ is continuous if and only if for any $U\in T_Y$, $f^{-1}(U)\in T_X$. In the case that $X$ and $Y$ are exactly the same topological space, $f$ is alternatively called a continuous endomorphism.

Third, a topological space $(Z,T_Z)$ is called a perfect square topological space if and only if there exists another topological space $(X,T_X)$ such that $Z=X\times X$ and $T_Z$ is exactly the product topology consisting of arbitrary unions of finite intersections of Cartesian products of elements of $T_x$ (all viewed embedded in $Z$). 

Fourth, a continuous binary operation (over a topological space $(X,T)$ is a continuous function $\oplus: X\times X\rightarrow X$, where $X\times X$ is assumed to have the product topology.

Fifth, a topological group $(G,+,e)$ is a group which is at the same time a topological space such that the operations $+$ and $Addinv:G\rightarrow G$ (sending $x\rightarrow -x$) are continuous functions. If the continuity of the inverse function is not required, then $(G,+,e)$ is a quasi-topological group. 

For a more detailed reading of the former concepts the interested reader may consult \cite{fraleigh} and \cite{munkres}.

\section{Conceptual Generation of the Notion of Topological Group in terms of Formal Conceptual Blending and Metaphorical Reasoning}

Due to the fact that we want to construct artificial specifications of mathematical notions `from scratch', we will describe along with the central axioms of each of the concepts, the minimal set-theoretical information needed to be able to do robust conceptual operations with them. Moreover, we present all the pseudo-codes in the most natural and clear way possible, so that working mathematicians with little experience with CASL would understand the essentials features of the mathematical structures involved.

\subsection{Continuous Binary Operation}\hspace*{\fill} \\

In the following specifications we will generate the notion of continuous binary operation as the formal blend between the notions of continuous functions (between topological spaces) and perfect square topological space.  
We use extra constants for some sorts, denoted with an additional `prime' symbol (e.g. $A'$), due to the fact that we need to be able to manipulate each sort as a `set' as well. Similarly, we will define a new constant for `simulating' the Cartesian product of a set with itself, because CASL does not deal with Cartesian products between sorts as constants. The importance of this technical trick can be better appreciated after reading completely each of the specifications. 
Now, such technicalities did not appear so explicitly in daily mathematical research because our minds do conceptual identifications almost automatically.

\begin{lstlisting}
(*\bf spec*) ContFunc =

(*\bf sorts*)   Sets, A, TA, PA, B, TB, PB;
A, TA, PA, B, TB, PB < Sets;
TA < PA; TB < PB;

%% A = domain of the function, TA = topology of A, PA = powerset of A
%% B = codomain of the function, TB = topology of B, PB = powerset of B

(*\bf ops*)     EmpSet, A', TA', PA', B', TB', PB' : Sets;
__ inter __ : Sets (*$\times$*) Sets (*$\to$*) Sets
Uni__ : Sets (*$\to$*) Sets
f: A (*$\to$*) B
inversef: TB (*$\to$*) TA 

(*\bf preds*)   __ subset __ : Sets (*$\times$*) Sets
__ el __ : Sets (*$\times$*) Sets 

%% Definition of A, TA and PA
(*$\forall$*)x : Sets
.x (*$\in$*) A (*$\Leftrightarrow$*) x el A'
.x (*$\in$*) TA (*$\Leftrightarrow$*) x el TA'
.x (*$\in$*) PA (*$\Leftrightarrow$*) x el PA'

%% Definition of B, TB and PB
(*$\forall$*)x : Sets
.x (*$\in$*) B (*$\Leftrightarrow$*) x el B'
.x (*$\in$*) TB (*$\Leftrightarrow$*) x el TB'
.x (*$\in$*) PB (*$\Leftrightarrow$*) x el PB'

%% Definition of subset
(*$\forall$*)x, y, z : Sets
.x subset y (*$\Leftrightarrow$*) (z el x (*$\Rightarrow$*) z el y)

%% Definition of ops
(*$\forall$*)x : Sets
..(*$\lnot$*)(x el EmpSet)
.x el PA' (*$\Leftrightarrow$*) x subset A'  
.x el PB' (*$\Leftrightarrow$*) x subset B'  
.TA' subset PA'  
.TB' subset PB'  
(*$\forall$*)x, y, z : Sets
. x el y inter z (*$\Leftrightarrow$*) x el y (*$\wedge$*) x el z 
(*$\forall$*)x, y : Sets
.x el Uni y (*$\Leftrightarrow$*) (*$\exists$*) z : Sets. z el y (*$\wedge$*) x el z

%% Specific axioms for a A as topological space
.EmpSet el TA'
.A' el TA'
(*$\forall$*)x, y : TA. x inter y el TA'
(*$\forall$*)x : Sets. x subset TA' (*$\Rightarrow$*) Uni x el TA'

%% Specific axioms for a B as topological space
.EmpSet el TB'
.B' el TB'
(*$\forall$*)x, y : TB. x inter y el TB'
(*$\forall$*)x : Sets. x subset TB' (*$\Rightarrow$*) Uni x el TB'

%% Inverse image of a set under a function
(*$\forall$*)y : TB. (*$\forall$*)x : A. x el inversef(y) (*$\Leftrightarrow$*) f(x) el y 

%% Condition of continuity
(*$\forall$*)y : TB. inversef(y) el TA'   

(*\bf end*)



(*\bf spec*) PerfSqTopSp =


(*\bf sorts*)   Sets, X, TX, PX, XX, TXX, PXX;
X, TX, PX, XX, TXX, PXX < Sets;
TX < PX; TXX < PXX;

(*\bf ops*)     EmpSet, X', TX', PX', XX', TXX', PXX' : Sets;
__ inter __ : Sets (*$\times$*) Sets (*$\to$*) Sets
__ ordpair __ : Sets (*$\times$*) Sets-> Sets
__ prod __ : Sets (*$\times$*) Sets (*$\to$*) Sets
Uni__ : Sets (*$\to$*) Sets

(*\bf preds*)   __ el __ : Sets
__ subset __ : Sets (*$\times$*) Sets;

%% X' and so on simulate the sorts
(*$\forall$*)x : Sets
.x (*$\in$*) X (*$\Leftrightarrow$*) x el X'
.x (*$\in$*) TX (*$\Leftrightarrow$*) x el TX'
.x (*$\in$*) PX (*$\Leftrightarrow$*) x el PX'
.x (*$\in$*) XX (*$\Leftrightarrow$*) x el XX'
.x (*$\in$*) TXX (*$\Leftrightarrow$*) x el TXX'
.x (*$\in$*) PXX (*$\Leftrightarrow$*) x el PXX'

%% Definition of subset
(*$\forall$*)x, y, z : Sets
.x subset y (*$\Leftrightarrow$*) (z el x (*$\Rightarrow$*) z el y)

%% Definition of ops
(*$\forall$*)x : Sets
.(*$\lnot$*)( x el EmpSet)
.x el PX' (*$\Leftrightarrow$*) x subset X'  
.TX' subset PX'  
.x el PXX' (*$\Leftrightarrow$*) x subset XX'  
.TX' subset PX'  
.TXX' subset PXX'  
(*$\forall$*)x, y, z : Sets . x el y inter z (*$\Leftrightarrow$*) x el y (*$\wedge$*) x el z 
(*$\forall$*)x, y : Sets . x el Uni y (*$\Leftrightarrow$*) (*$\exists$*) z : Sets. z el y (*$\wedge$*) x el z

%% Defining ordpair
(*$\forall$*)x, y, z : Sets. z el x ordpair y (*$\Leftrightarrow$*) (*$\forall$*)s : Sets. (s el z (*$\Leftrightarrow$*) s = x) (*$\lor$*) (s el z (*$\Leftrightarrow$*) s = x (*$\lor$*) s = y)  

%% Defining prod
(*$\forall$*)A, B, z : Sets . z el A prod B (*$\Leftrightarrow$*) (*$\exists$*) x, y : Sets . x el A (*$\wedge$*) y el B (*$\wedge$*) z = x ordpair y 	

%%Specific axioms for a X as topological space
.EmpSet el TX'
.X' el TX'
(*$\forall$*)x, y : TX. x inter y el TX'
(*$\forall$*)x : Sets. x subset TX' (*$\Rightarrow$*) Uni x el TX'


%%Specific axioms for XX as topological space
.EmpSet el TXX'
.XX' el TXX'
(*$\forall$*)x, y : TXX. x inter y el TXX'
(*$\forall$*)x : Sets. x subset TXX' (*$\Rightarrow$*) Uni x el TXX'

%% XX' is product and TXX' is the product topology
. XX' = X' prod X'
(*$\forall$*)z: Sets . z el TXX' <=> (*$\forall$*)w: Sets .(w el z => exists x, y : TX . w el x prod y /\  x prod y subset z) 

(*\bf end*)


(*\bf spec*) Generic =

(*\bf sorts*)   Sets, X, XX, TX, TXX, PX, PXX;

ops 	EmpSet, X', XX', TX', TXX', PX', PXX' : Sets
__ inter __ : Sets (*$\times$*) Sets (*$\to$*) Sets
Uni__ : Sets (*$\to$*) Sets

(*\bf preds*)   __ el __ : Sets (*$\times$*) Sets;
__ subset __ : Sets (*$\times$*) Sets;

(*\bf end*)


view I1 : Generic to PerfSqTopSp =

Sets (*$\mapsto$*) Sets, X (*$\mapsto$*) X, XX (*$\mapsto$*) XX, TX (*$\mapsto$*) TX, TXX (*$\mapsto$*) TXX, PX (*$\mapsto$*) PX, PXX (*$\mapsto$*) PXX, X' (*$\mapsto$*) X', XX' (*$\mapsto$*) XX', TX' (*$\mapsto$*) TX', TXX' (*$\mapsto$*) TXX', PX' (*$\mapsto$*) PX', PXX' (*$\mapsto$*) PXX', EmpSet (*$\mapsto$*) EmpSet, __ el __ (*$\mapsto$*) __ el __, __ subset __ (*$\mapsto$*) __ subset __, __ inter __ (*$\mapsto$*) __ inter__, Uni__ (*$\mapsto$*) Uni__

(*\bf end*)


view I2 : Generic to ContFunc =

Sets (*$\mapsto$*) Sets, X (*$\mapsto$*) B, XX (*$\mapsto$*) A, TX (*$\mapsto$*) TB, TXX (*$\mapsto$*) TA, PX (*$\mapsto$*) PB, PXX (*$\mapsto$*) PA, X' (*$\mapsto$*) B', XX' (*$\mapsto$*) A', TX' (*$\mapsto$*) TB', TXX' (*$\mapsto$*) TA', PX' (*$\mapsto$*) PB', PXX' (*$\mapsto$*) PA', EmpSet (*$\mapsto$*) EmpSet, __ el __ (*$\mapsto$*) __ el __, __ subset __ (*$\mapsto$*) __ subset__, __ inter __ (*$\mapsto$*) __ inter __, Uni__ (*$\mapsto$*) Uni__

(*\bf end*)

(*\bf spec*) Colimit = combine I1, I2

\end{lstlisting}

By computing the corresponding blend (i.e. colimit), we obtain a specification of the notion of continuous binary operation. So, after doing some improvements in the presentation (e.g. updating names of sorts), one essentially obtains an specification like the following: 

\begin{lstlisting}
	(*\bf spec*) contBinFunc =
	
	(*\bf sorts*) PX, PXX, Sets, TX, TXX, X, XX
	(*\bf sorts*) TX < PX; TXX < PXX; PX, PXX, X, XX < Sets
	
	(*\bf op*) EmpSet : Sets
	(*\bf op*) PX' : Sets
	(*\bf op*) PXX' : Sets
	(*\bf op*) TX' : Sets
	(*\bf op*) TXX' : Sets
	(*\bf op*) Uni__ : Sets (*$\to$*) Sets
	(*\bf op*) X' : Sets
	(*\bf op*) XX' : Sets
	(*\bf op*) __inter__ : Sets (*$\times$*) Sets (*$\to$*) Sets
	(*\bf op*) __ordpair__ : Sets (*$\times$*) Sets (*$\to$*) Sets
	(*\bf op*) __prod__ : Sets (*$\times$*) Sets (*$\to$*) Sets
	(*\bf op*) f : XX (*$\to$*) X
	(*\bf op*) inversef : TX (*$\to$*) TXX
	
	(*\bf pred*) __el__ : Sets (*$\times$*) Sets
	(*\bf pred*) __subset__ : Sets (*$\times$*) Sets
	
	%% Simulation of X, TX and PX
	(*$\forall$*)x : Sets . x (*$\in$*) X (*$\Leftrightarrow$*) x el X' %(Ax1)%
	(*$\forall$*)x : Sets . x (*$\in$*) TX (*$\Leftrightarrow$*) x el TX' %(Ax2)%
	(*$\forall$*)x : Sets . x (*$\in$*) PX (*$\Leftrightarrow$*) x el PX' %(Ax3)%
	
	%% Simulation of XX, TXX, PXX
	(*$\forall$*)x : Sets . x (*$\in$*) XX (*$\Leftrightarrow$*) x el XX' %(Ax1_25)%
	(*$\forall$*)x : Sets . x (*$\in$*) TXX (*$\Leftrightarrow$*) x el TXX' %(Ax2_36)%
	(*$\forall$*)x : Sets . x (*$\in$*) PXX (*$\Leftrightarrow$*) x el PXX' %(Ax3_42)%
	
	%% Definition of subset
	(*$\forall$*)x, y, z : Sets . x subset y (*$\Leftrightarrow$*) (z el x (*$\Rightarrow$*) z el y) %(Ax7)%
	
	%% Definition of ops
	. TXX' subset PXX' %(Ax11_27)%
	. TX' subset PX' %(Ax12_28)%
	(*$\forall$*)x : Sets . not x el EmpSet %(Ax8)%
	(*$\forall$*)x : Sets . x el PXX' (*$\Leftrightarrow$*) x subset XX' %(Ax9_48)%
	(*$\forall$*)x : Sets . x el PX' (*$\Leftrightarrow$*) x subset X' %(Ax10_26)%
	
	(*$\forall$*)x, y, z : Sets . x el y inter z (*$\Leftrightarrow$*) x el y (*$\wedge$*) x el z %(Ax11)%
	(*$\forall$*)x, y : Sets. x el Uni y (*$\Leftrightarrow$*) (*$\exists$*) z : Sets . z el y (*$\wedge$*) x el z %(Ax12)%
	(*$\forall$*)x, y, z : Sets. z el x ordpair y (*$\Leftrightarrow$*) (*$\forall$*)s : Sets. (s el z (*$\Leftrightarrow$*) s = x) (*$\lor$*) (s el z (*$\Leftrightarrow$*) s = x (*$\lor$*) s = y)%(Ax13)%
	(*$\forall$*)A, B, z : Sets . z el A prod B (*$\Leftrightarrow$*) (*$\exists$*) x, y : Sets . x el A (*$\wedge$*) y el B (*$\wedge$*) z = x ordpair y %(Ax14)%
	
	%% XX, TXX is product
	. XX' = X' prod X' %(Ax23)%
	(*$\forall$*)z: Sets . z el TXX' <=> (*$\forall$*)w: Sets .(w el z => exists x, y : TX . w el x prod y /\  x prod y subset z) %(Ax24)%
	
	%% TX is topology
	. EmpSet el TX' %(Ax15)%
	. X' el TX' %(Ax16)%
	(*$\forall$*)x, y : TX . x inter y el TX' %(Ax17)%
	(*$\forall$*)x : Sets . x subset TX' (*$\Rightarrow$*) Uni x el TX' %(Ax18)%
	
	%% TXX is topology
	. EmpSet el TXX' %(Ax15_31)%
	. XX' el TXX' %(Ax16_32)%
	(*$\forall$*)x, y : TXX . x inter y el TXX' %(Ax17_33)%
	(*$\forall$*)x : Sets . x subset TXX' (*$\Rightarrow$*) Uni x el TXX' %(Ax18_34)%
	
	%% Definition of inversef
	(*$\forall$*)y : TX; x : XX . x el inversef(y) (*$\Leftrightarrow$*) f(x) el y %(Ax23_40)%
	
	%% f is continuous
	(*$\forall$*)y : TX . inversef(y) el TXX' %(Ax24_41)%
	
	(*\bf end*)

\end{lstlisting}

\subsection{Quasi-topological Groups}\hspace*{\fill} \\

Let us combine the latter blended concept (i.e. continuous binary operations) with (an enriched form of) the notion of group to generate the concept of quasi-topological groups.\footnote{Due to space constraits the reader can see the whole specification in the following github file: \url{https://github.com/yoeherrera/pseudocode-for-topological-groups-in-CASL/blob/main/4-quasitopological\%20group.tex}}

\subsection{Continuous Endomorphisms}\hspace*{\fill} \\

We will obtain the notion of continuous endomorphism starting with continuous functions (between topological spaces) and doing a conceptual identification between the domain and the codomain of the corresponding map. Explicitly, in the former specification of the conceptual space of continuous functions, we declare the equality of the corresponding sorts of the domain and codomain as follows: $A\cong B; TA\cong TB$ and $PA \cong PB$. In this way, we obtain the concrete specification of the notion of continuous endomorphism.\footnote{Again, the reader can find the explicit especification in the following github repository \url{https://github.com/yoeherrera/pseudocode-for-topological-groups-in-CASL/blob/main/3-continuous\%20endomorphism.tex}}



\subsection{Topological Groups}\hspace*{\fill} \\

Finally, we generate the concept of Topological Group as the following blend (i.e. colimit) of the former two (specifications of) concepts; i.e., quasi-topological groups and continuous endomorphisms:

\begin{lstlisting}
(*\bf spec*) QuasiTopGroup =

(*\bf sorts*) X, XX, PX, PXX, Sets, TX, TXX
(*\bf sorts*) TX < PX; TXX < PXX; X, XX, PX, PXX < Sets

(*\bf op*) EmpSet : Sets
(*\bf op*) PX' : Sets
(*\bf op*) PXX' : Sets
(*\bf op*) TX' : Sets
(*\bf op*) TXX' : Sets
(*\bf op*) X' : Sets
(*\bf op*) XX' : Sets

(*\bf op*) __ordpair__ : Sets (*$\times$*) Sets (*$\to$*) Sets
(*\bf op*) __Xpair__ : X (*$\times$*) X (*$\to$*) XX
(*\bf op*) embedding : X (*$\to$*) Sets

(*\bf op*) Uni__ : Sets (*$\to$*) Sets
(*\bf op*) __inter__ : Sets (*$\times$*) Sets (*$\to$*) Sets
(*\bf op*) __prod__ : Sets (*$\times$*) Sets (*$\to$*) Sets

(*\bf op*) 0 : X
(*\bf op*) Addinv : X (*$\to$*) X
(*\bf op*) __+__ : X (*$\times$*) X (*$\to$*) X
(*\bf op*) inverseplus : TX (*$\to$*) TXX %% inverse of ++
(*\bf op*) ++ : XX (*$\to$*) X


(*\bf pred*) __el__ : Sets (*$\times$*) Sets
(*\bf pred*) __subset__ : Sets (*$\times$*) Sets


%% Simulation
(*$\forall$*)x : Sets . x (*$\in$*) X (*$\Leftrightarrow$*) x el X' 
(*$\forall$*)x : Sets . x (*$\in$*) TX (*$\Leftrightarrow$*) x el TX'
(*$\forall$*)x : Sets . x (*$\in$*) PX (*$\Leftrightarrow$*) x el PX'
(*$\forall$*)x : Sets . x (*$\in$*) XX (*$\Leftrightarrow$*) x el XX'
(*$\forall$*)x : Sets . x (*$\in$*) TXX (*$\Leftrightarrow$*) x el TXX' 
(*$\forall$*)x : Sets . x (*$\in$*) PXX (*$\Leftrightarrow$*) x el PXX' 

%% Definition of TXX', TX', PX', PXX' and EmpSet
. TXX' subset PXX' 
. TX' subset PX' 
(*$\forall$*)x : Sets . x el PXX' (*$\Leftrightarrow$*) x subset XX' 
(*$\forall$*)x : Sets . x el PX' (*$\Leftrightarrow$*) x subset X' 
(*$\forall$*)x : Sets . not x el EmpSet 

%% Definition of subset
(*$\forall$*)x, y, z : Sets . x subset y (*$\Leftrightarrow$*) (z el x (*$\Rightarrow$*) z el y) 

%% Definition of ordpair, embedding and Xpair
(*$\forall$*)x, y, z : Sets . z el x ordpair y (*$\Leftrightarrow$*) (*$\forall$*)s : Sets . (s el z (*$\Leftrightarrow$*) s = x) (*$\lor$*) (s el z (*$\Leftrightarrow$*) s = x (*$\lor$*) s = y) %(Ax13)%
(*$\forall$*)x : X . x = embedding(x) %(Ax3)%
(*$\forall$*)a, b : X . a Xpair b = embedding(a) ordpair embedding(b)

%% Definition of Uni, inter and prod
(*$\forall$*)x, y, z : Sets . x el y inter z (*$\Leftrightarrow$*) x el y (*$\wedge$*) x el z %(Ax11)%
(*$\forall$*)x, y : Sets . x el Uni y (*$\Leftrightarrow$*) (*$\exists$*) z : Sets . z el y (*$\wedge$*) x el z %(Ax12)%
(*$\forall$*)A, B, z : Sets . z el A prod B (*$\Leftrightarrow$*) (*$\exists$*) x, y : Sets . x el A (*$\wedge$*) y el B (*$\wedge$*) z = x ordpair y 

%% Group axioms
(*$\forall$*)x, y, z : X . (x + y) + z = x + (y + z) 
(*$\forall$*)x : X . x + 0 = x 
(*$\forall$*)x : X . Addinv(x) + x = 0

%% TX' is topology
. EmpSet el TX' 
. X' el TX' 
(*$\forall$*)x, y : TX . x inter y el TX'
(*$\forall$*)x : Sets . x subset TX' (*$\Rightarrow$*) Uni x el TX' 

%% TXX' is topology
. EmpSet el TXX' 
. XX' el TXX' 
(*$\forall$*)x, y : TXX . x inter y el TXX'
(*$\forall$*)x : Sets . x subset TXX' (*$\Rightarrow$*) Uni x el TXX'

%% XX' are products
. XX' = X' prod X'
(*$\forall$*)z : Sets . z el TXX' (*$\Leftrightarrow$*) (*$\exists$*) x, y : TX . z = x prod y

%% Definition of ++ and inversef
(*$\forall$*)a, b : X . a + b = ++(a Xpair b) 
(*$\forall$*)y : TX; x : XX . x el inverseplus(y) (*$\Leftrightarrow$*) ++(x) el y 

%% ++ is continuous
(*$\forall$*)y : TX . inverseplus(y) el TXX'

(*\bf end*)


(*\bf spec*) ContEndo = 

(*\bf sorts*)  Sets, A, TA, PA;
A, TA, PA < Sets;
TA < PA;

(*\bf ops*) EmpSet, A', TA', PA' : Sets;
__inter__ : Sets (*$\times$*) Sets (*$\to$*) Sets
Uni__ : Sets (*$\to$*) Sets
Addinv: A (*$\to$*) A
inverseAddinv: TA (*$\to$*) TA %% inverse of f

(*\bf preds*) __subset__ : Sets (*$\times$*) Sets
__el__ : Sets (*$\times$*) Sets 

%% Definition of A, TA and PA
(*$\forall$*)x : Sets
.x (*$\in$*) A (*$\Leftrightarrow$*) x el A'
.x (*$\in$*) TA (*$\Leftrightarrow$*) x el TA'
.x (*$\in$*) PA (*$\Leftrightarrow$*) x el PA'

%% Definition of subset
(*$\forall$*)x, y, z : Sets
.x subset y (*$\Leftrightarrow$*) (z el x (*$\Rightarrow$*) z el y)

%% Definition of ops
(*$\forall$*)x : Sets
..(*$\lnot$*)( x el EmpSet)
.x el PA' (*$\Leftrightarrow$*) x subset A'
.TA' subset PA'
(*$\forall$*)x, y, z : Sets
. x el y inter z (*$\Leftrightarrow$*) x el y (*$\wedge$*) x el z
(*$\forall$*)x, y : Sets
.x el Uni y (*$\Leftrightarrow$*) (*$\exists$*) z : Sets. z el y (*$\wedge$*) x el z

%% Specific axioms for A as topological space
.EmpSet el TA'
.A' el TA'
(*$\forall$*)x, y : TA. x inter y el TA'
(*$\forall$*)x : Sets. x subset TA' (*$\Rightarrow$*) Uni x el TA'

%% Inverse image of a set under a function
(*$\forall$*)y : TA. (*$\forall$*)x : A. x el inverseAddinv(y) (*$\Leftrightarrow$*) Addinv(x) el y

%% Condition of continuity
(*$\forall$*)y : TA. inverseAddinv(y) el TA'

(*\bf end*)


(*\bf spec*) Generic = 

(*\bf sorts*)  Sets, X, TX, PX

ops 	EmpSet, X', TX', PX' : Sets
__inter__ : Sets (*$\times$*) Sets (*$\to$*) Sets
Uni__ : Sets (*$\to$*) Sets
Addinv : X (*$\to$*) X

(*\bf preds*) __el__ : Sets (*$\times$*) Sets
__subset__ : Sets (*$\times$*) Sets

(*\bf end*)


view I1 : Generic to QuasiTopGroup =

Sets (*$\mapsto$*) Sets, X (*$\mapsto$*) X, TX (*$\mapsto$*) TX, PX (*$\mapsto$*) PX, 
EmpSet (*$\mapsto$*) EmpSet, X' (*$\mapsto$*) X', TX' (*$\mapsto$*) TX', PX' (*$\mapsto$*) PX', 
__inter__ (*$\mapsto$*) __inter__, Uni__ (*$\mapsto$*) Uni__, Addinv (*$\mapsto$*) Addinv, 
__el__ (*$\mapsto$*) __el__, __subset__ (*$\mapsto$*) __subset__

(*\bf end*)


view I2 : Generic to ContEndo =

Sets (*$\mapsto$*) Sets, X (*$\mapsto$*) A, TX (*$\mapsto$*) TA, PX (*$\mapsto$*) PA,
EmpSet (*$\mapsto$*) EmpSet, X' (*$\mapsto$*) A', TX' (*$\mapsto$*) TA', PX' (*$\mapsto$*) PA',
__inter__ (*$\mapsto$*) __inter__, Uni__ (*$\mapsto$*) Uni__, Addinv (*$\mapsto$*) Addinv,
__el__ (*$\mapsto$*) __el__, __subset__ (*$\mapsto$*) __subset__

(*\bf end*)


(*\bf spec*) TopGroup = combine I1, I2
\end{lstlisting}

After doing the computation of the colimit we essentially obtain the classic concept of topological group:

\begin{lstlisting}
(*\bf sorts*)  PX, PXX, Sets, TX, TXX, X, XX
(*\bf sorts*)  TX < PX; TXX < PXX; PX, PXX, X, XX < Sets

(*\bf op*) 	0 : X
(*\bf ops*) EmpSet, X', XX', TX', TXX', PX', PXX' : Sets
(*\bf op*) 	++ : XX (*$\to$*) X
(*\bf op*) 	Uni__ : Sets (*$\to$*) Sets
(*\bf op*) 	__inter__ : Sets (*$\times$*) Sets (*$\to$*) Sets
(*\bf op*) 	__Xpair__ : X (*$\times$*) X (*$\to$*) XX
(*\bf op*) 	__ordpair__ : Sets (*$\times$*) Sets (*$\to$*) Sets
(*\bf op*) 	__prod__ : Sets (*$\times$*) Sets (*$\to$*) Sets
(*\bf op*) 	embedding : X (*$\to$*) Sets
(*\bf op*) 	Addinv : X (*$\to$*) X
(*\bf op*) 	__+__ : X (*$\times$*) X (*$\to$*) X
(*\bf op*) 	inverseAddinv : TX (*$\to$*) TX
(*\bf op*) 	inverseplus : TX (*$\to$*) TXX'

(*\bf pred*) 	__el__ : Sets (*$\times$*) Sets
(*\bf pred*) 	__subset__ : Sets (*$\times$*) Sets

%% Simulation
(*$\forall$*)x : Sets . x (*$\in$*) X (*$\Leftrightarrow$*) x el X' %(Ax1)%
(*$\forall$*)x : Sets . x (*$\in$*) TX (*$\Leftrightarrow$*) x el TX' %(Ax2)%
(*$\forall$*)x : Sets . x (*$\in$*) PX (*$\Leftrightarrow$*) x el PX' %(Ax3_13)%
(*$\forall$*)x : Sets . x (*$\in$*) XX (*$\Leftrightarrow$*) x el XX' %(Ax1_25)%
(*$\forall$*)x : Sets . x (*$\in$*) TXX (*$\Leftrightarrow$*) x el TXX' %(Ax2_36)%
(*$\forall$*)x : Sets . x (*$\in$*) PXX (*$\Leftrightarrow$*) x el PXX' %(Ax3_42)%

. TXX' subset PXX' %(Ax11_27)%
. TX' subset PX' %(Ax12_28)%
(*$\forall$*)x : Sets . x el PXX' (*$\Leftrightarrow$*) x subset XX' %(Ax9_48)%
(*$\forall$*)x : Sets . x el PX' (*$\Leftrightarrow$*) x subset X' %(Ax10_26)%
(*$\forall$*)x : Sets . not x el EmpSet %(Ax5_45)%

%% Definition of subset
(*$\forall$*)x, y, z : Sets . x subset y (*$\Leftrightarrow$*) z el x (*$\Rightarrow$*) z el y %(Ax4_44)%

%% Definition of ordpair, embedding and Xpair
(*$\forall$*)x, y, z : Sets . z el x ordpair y<=> (*$\forall$*)s : Sets . (s el z (*$\Leftrightarrow$*) s = x) (*$\lor$*) (s el z (*$\Leftrightarrow$*) s = x (*$\lor$*) s = y) %(Ax13)%
(*$\forall$*)x : X . x = embedding(x) %(Ax3)%
(*$\forall$*)a, b : X . a Xpair b = embedding(a) ordpair embedding(b) %(Ax15)%

%% Definition of Uni, inter and prod
(*$\forall$*)x, y, z : Sets . x el y inter z (*$\Leftrightarrow$*) x el y (*$\wedge$*) x el z %(Ax11)%
(*$\forall$*)x, y : Sets . x el Uni y (*$\Leftrightarrow$*) (*$\exists$*) z : Sets . z el y (*$\wedge$*) x el z %(Ax12)%
(*$\forall$*)A, B, z : Sets . z el A prod B (*$\Leftrightarrow$*) (*$\exists$*) x, y : Sets . x el A (*$\wedge$*) y el B (*$\wedge$*) z = x ordpair y %(Ax14)%

%% Group axioms
(*$\forall$*)x, y, z : X . (x + y) + z = x + (y + z) %(Ax4)%
(*$\forall$*)x : X . x + 0 = x %(Ax5)%
(*$\forall$*)x : X . Addinv(x) + x = 0 %(Ax6)%

%% TX is topology
. EmpSet el TX' %(Ax10_36)%
. X' el TX' %(Ax11_37)%
(*$\forall$*)x, y : TX . x inter y el TX' %(Ax12_38)%
(*$\forall$*)x : Sets . x subset TX' (*$\Rightarrow$*) Uni x el TX' %(Ax13_39)%

%% TXX is topology
. EmpSet el TXX' %(Ax15_31)%
. XX' el TXX' %(Ax16_32)%
(*$\forall$*)x, y : TXX . x inter y el TXX' %(Ax17_33)%
(*$\forall$*)x : Sets . x subset TXX' (*$\Rightarrow$*) Uni x el TXX' %(Ax18_34)%

%% XX' is product and TXX' is product topology
. XX' = X' prod X' %(Ax23)%
(*$\forall$*)z: Sets . z el TXX' <=> (*$\forall$*)w: Sets .(w el z => exists x, y : TX . w el x prod y /\  x prod y subset z) %(Ax24)%

%% Definition of ++, inverseplus and inverseAddinv
(*$\forall$*)a, b : X . a + b = ++(a Xpair b) %(Ax10)%
(*$\forall$*)y : TX; x : XX . x el inverseplus(y) (*$\Leftrightarrow$*) ++(x) el y %(Ax23_40)%
(*$\forall$*)y : TX; x : X . x el inverseAddinv(y) (*$\Leftrightarrow$*) Addinv(x) el y %(Ax14_40)%

%% Condition of continuity
(*$\forall$*)y : TX . inverseplus(y) el TXX' %(Ax24_41)%
(*$\forall$*)y : TX . inverseAddinv(y) el TX' %(Ax15_41)%
\end{lstlisting}

The former specification is considerably larger than the one usually given in the text books due to the fact that we include additionally the minimal set-theoretical information required to define an essentially autonomous concept, which can be coherently described with the semantic tools of CASL and HETS. In fact, the former conceptual computations were explicitly run and proved in HETS.

\section{Summarized Results}

We also can state the results presented in the former specifications in the form of a global (meta-)theorem describing categorical constructions done before in terms of colimits:
\begin{theorem}
The concept of topological group, viewed as a unique object of the co-complete category of many-sorted first-order theories with axiom-preserving signature morphisms, can be generated recursively by means of three formal colimits (blends), starting from the concepts of enriched groups, continuous functions and continuous endomorphisms.
\end{theorem}

\section{Discussion and Conclusions}

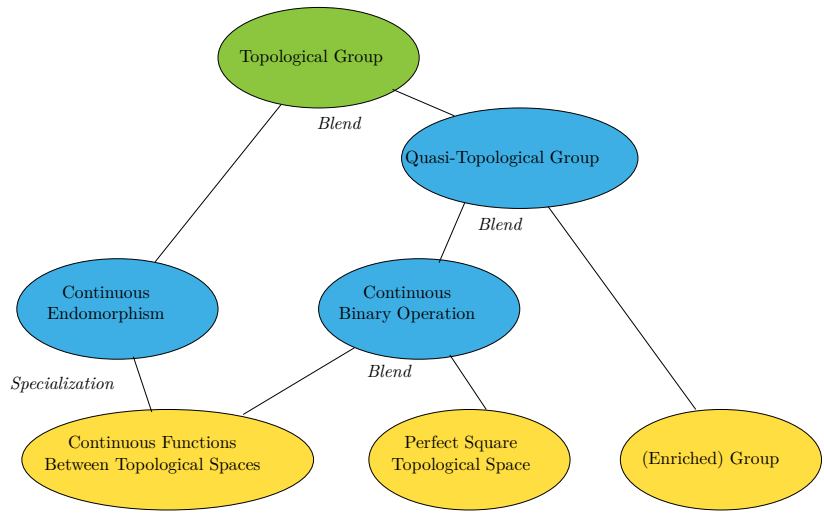
\begin{figure}[htpb]\label{fig:diag1}
\centering
    \resizebox{0.8\textwidth}{!}{%
\begin{tikzpicture}
    \draw[fill=LimeGreen] (0,0) ellipse (2cm and 1cm);
    \draw[fill=CornflowerBlue] (4,-2) ellipse (2.35 cm and 1cm);
    \draw[fill=Goldenrod] (8,-8) ellipse (2cm and 1cm);
    \draw[fill=CornflowerBlue] (2,-5) ellipse (2cm and 1cm);
    \draw[fill=CornflowerBlue] (-4,-5) ellipse (2cm and 1cm);
    \draw[fill=Goldenrod] (3,-8) ellipse (2cm and 1cm);
    \draw[fill=Goldenrod] (-3,-8) ellipse (2.9cm and 1cm);
    
    \coordinate (A) at (-1.7,0);
    \coordinate (B) at (1.6,-2);
    \coordinate (C) at (6.3,-8);
    \coordinate (D) at (0.1,-4.9);
    \coordinate (E) at (-5.7,-4.9);
    \coordinate (F) at (1.15,-7.9);
    \coordinate (G) at (-5.75,-7.9);

    \node[right] at (A) {Topological Group};
    \node[right] at (B) {Quasi-Topological Group};
    \node[right] at (C) {(Enriched) Group};
    \node[right] at (D) {\begin{tabular}{c} Continuous \\ Binary Operation \end{tabular}};
    \node[right] at (E)  {\begin{tabular}{c} Continuous \\ Endomorphism \end{tabular}} {};
    \node[right] at (F) {\begin{tabular}{c} Perfect Square  \\ Topological Space \end{tabular}};
    \node[right] at (G) {\begin{tabular}{c} Continuous Functions\\ Between Topological Spaces\end{tabular}};
    
   \coordinate (H) at (0,0);
    \coordinate (I) at (3.5,-1.5);
    \coordinate (J) at (-4,-5);
    \coordinate (K) at (2,-5);
    \coordinate (L) at (7.5,-7);
    \coordinate (M) at (4,-8);
    \coordinate (N) at (-3,-8);
    
      \draw[shorten <=1.6cm,shorten >=0.87cm] (H) -- (I);
     \draw[shorten <=1.2cm,shorten >=1.2cm] (H) -- (J);
     \draw[shorten <=1.5cm,shorten >=1cm] (I) -- (K);
     \draw[shorten <= 1.8cm] (I) -- (L);
     \draw[shorten <=1.1cm,shorten >=1.2cm] (K) -- (M);
     \draw[shorten <=1.5cm,shorten >=1.75cm] (K) -- (N);
     \draw[shorten <=1.0cm,shorten >=1cm] (J) -- (N);
    
    \coordinate (O) at (0.4,-1.3);
    \coordinate (P) at (3.6,-3.3);
    \coordinate (Q) at (-5.1,-6.5);
    \coordinate (R) at (1.4,-6.24);
    
    \node at (O) {{\it Blend}};
    \node at (P) {{\it Blend}};
    \node at (Q) {{\it Specialization}};
    \node at (R) {{\it Blend}};
\label{fig:1}
    
\end{tikzpicture}}%

\caption{Diagrammatic Representation for the recursive generation of the concept of Topological
Group through Formal Conceptual Blending and specialization}
\end{figure}

In Figure \ref{fig:1}, we present a diagrammatic summary of the whole recursive generation done through formal conceptual blending with the help of conceptual identification. 

The fact that we explicitly find artificial specifications, or pseudo-pre-code, of sophisticated concepts in abstract algebra and topology represents valuable domain-specific evidence for the universality of the meta-tools described by means of categorical formalizations of conceptual blending (and, in an indirect way, by the more informal categorical approach of conceptual identification made in terms of sorts' identifications). 

The former results also promote the thesis that the potential scope of the co-creative power of artificial interactive systems regarding mathematical invention goes beyond the typical elementary structures classically studied, e.g., the complex numbers \cite{fleuriot}. 
Moreover, this artificial meta-generation represents additional evidence of the formal soundness and strength of artificial mathematical intelligence regarding (artificial) conceptual generation in advanced mathematics and ZFC Set Theory \cite[Ch.3]{mendelsonlogic}.

Finally, this research goes towards the development of new forms of conceptual co-creative cybernetics in the domain of interactive mathematical creation. In fact, previous forms of this new kind of cybernetics were developed within the multidisciplinary research consortium COINVENT \cite{coinvent}, \cite{cinventionbook}. Specifically, in \cite{conceptinvention3}, an interactive co-creative computational prototype called COBBLE is presented, materializing artificial co-innovative reasoning in mathematics and music (harmonization) based on notions coming not only from conceptual blending theory, but also from analogical reasoning, formal ontology theory, logic programming and formal methods. So, the collection of results presented here can be seen as a first theoretical step towards extensions of such computational prototypes to broader mathematically-based disciplines. 

\section*{Acknowledgements}
The second author thank the Universidad Autonoma de Bucaramanga for all the support (Grant I56073) and the Instituci\'on Universitaria Tecnol\'ogico de Antioquia. Danny A. J. Gomez-Ramirez thanks Fabian Suarez for all the sincere friendship and support.

\providecommand{\bysame}{\leavevmode\hbox to3em{\hrulefill}\thinspace}
\providecommand{\MR}{\relax\ifhmode\unskip\space\fi MR }
\providecommand{\MRhref}[2]{%
  \href{http://www.ams.org/mathscinet-getitem?mr=#1}{#2}
}
\providecommand{\href}[2]{#2}

\end{document}